\documentclass[letterpaper]{article} 
\usepackage[preprint]{arxiv}  
\usepackage[hyphens]{url}  
\usepackage{graphicx} 
\usepackage{natbib}  
\usepackage{caption} 
\usepackage{algorithm}
\usepackage{algorithmic}

\usepackage{booktabs}
\usepackage{multirow}   
\usepackage{xcolor}     
\usepackage{tabularx}   
\usepackage[dvipsnames]{xcolor}
\usepackage{amsmath}
\usepackage{siunitx}
\usepackage{arydshln}
\usepackage{microtype}
\usepackage{amssymb}

\newcommand{\down}[1]{{\scriptsize $\downarrow$#1}}
\newcommand{\phspace}[1]{\phantom{\down{#1}}}

\usepackage{newfloat}
\usepackage{listings}
\DeclareCaptionStyle{ruled}{labelfont=normalfont,labelsep=colon,strut=off} 
\floatstyle{ruled}
\newfloat{listing}{tb}{lst}{}
\floatname{listing}{Listing}

\usepackage{booktabs}

\usepackage[most]{tcolorbox}
\usepackage{fvextra}
\usepackage{caption}

\newtcolorbox{promptbox}[1]{
  breakable, enhanced,
  colback=gray!8, colframe=black!70, coltitle=white,   
  fonttitle=\bfseries\small, title={#1},
  boxrule=1pt, arc=2pt,                             
  left=4pt, right=4pt, top=3pt, bottom=3pt,
  before skip=6pt, after skip=6pt
}

\title{CRISP: Critical Step Perception for Training Efficient Deep Search Agents}
\author{
    Haosi Mo\textsuperscript{\rm 1,\rm 2}\equalcontrib,
    Zihao Yan\textsuperscript{\rm 1,\rm 2}\equalcontrib,
    Ruiqing Zhang\textsuperscript{\rm 1}\corresponding,
    Zhongli Li\textsuperscript{\rm 1},
    Hexuan Deng\textsuperscript{\rm 2},
    Xuebo Liu\textsuperscript{\rm 2}\corresponding,
    Min Zhang\textsuperscript{\rm 2}
}
\affiliations{
    \textsuperscript{\rm 1}Baidu Inc., China\\
    \textsuperscript{\rm 2}Institute of Computing and Intelligence, Harbin Institute of Technology, Shenzhen, China\\
    zhangruiqing01@baidu.com, liuxuebo@hit.edu.cn
}

\begin{document}

\maketitle

\begin{abstract}
Large language models (LLMs) are increasingly extended into deep search agents that solve complex questions through multi-step interaction with external search and browsing tools. However, existing agents often incur substantial computational and interaction costs, generating lengthy trajectories 
that contain redundant queries, inefficient exploration, and irrelevant observations. Existing efficiency-oriented methods usually encourage agents to use tools less frequently, but treating all tool interactions uniformly may also suppress steps that gather necessary evidence. In this paper, we propose CRISP, a framework for training efficient deep search agents through critical step perception. Unlike prior efficiency methods that uniformly penalize tool use, CRISP distinguishes interactions that gather necessary evidence from redundant ones and shapes the training reward to preserve the former while pruning the latter, improving efficiency without sacrificing the evidence needed for correct answers. Specifically, CRISP first constructs critical-step labels with Backward Evidence Induction: starting from the final answer, a strong model traverses a completed search trajectory backward and judges whether each tool-interaction step provides or preserves evidence for the final answer. We then distill these step-wise judgments into a smaller critical-step recognizer, enabling full-trajectory analysis in a single pass. During policy optimization, an efficiency-aware reward is applied only to successful rollouts. Experiments on BrowseComp and HLE-Verified show that CRISP maintains competitive final-answer accuracy while reducing average interaction turns by 15.1\% and 33.2\%, respectively, demonstrating substantial improvements in interaction efficiency.
\end{abstract}


\section{Introduction}

Large language models (LLMs) are increasingly extended into interactive search agents that acquire and synthesize external information~\citep{xi2025survey,huang2025deep}. In complex scenarios, answering questions often requires multi-step evidence gathering and reasoning rather than a single retrieval or one-shot response~\citep{gupta2026deepsearchqa,lan2025deepwidesearch,wei2025browsecomp}. This has motivated deep search agents that combine LLM reasoning with external search, browsing, and document-reading tools to tackle challenging problems~\citep{huang-etal-2025-manusearch,jin2025searchr}. As these agents tackle increasingly difficult tasks, their full interaction trajectories provide a rich record of how models seek, access, and use evidence to support the final answer.

Despite their growing capability, deep search agents often incur high interaction costs on complex tasks. Agentic reinforcement learning can induce redundant tool use, encouraging over-reliance on external tools rather than selective interaction~\citep{chen2026efficient}. This issue is also evident in representative open-source agents. As shown in Table~\ref{tab:bc_interaction_costs}, they still average over 30 interaction turns on BrowseComp~\citep{wei2025browsecomp} under the same max\_turn $=60$ budget. The deeper problem, however, is not trajectory length itself, but that long traces mix necessary evidence gathering with repeated queries, weak clues, inefficient exploration, and observations unused in the final answer. Such redundancy wastes computation and thinking effort while increasing the burden of long-context understanding and critical information discrimination.

\begin{table}[t]
    \centering
    \small
    \caption{Representative open-source agents still require many interaction turns on BrowseComp.}
    \label{tab:bc_interaction_costs}
    \setlength{\tabcolsep}{4pt}
        \begin{tabular}{lcc}
        \toprule
        \textbf{Model} & \textbf{Acc. (\%)} & \textbf{Avg. Turns} \\
        \midrule
        GLM-4.5-Air~\citep{5team2025glm45agenticreasoningcoding} 
        & 14.49 & 31.23 \\
        GLM-4.7~\citep{glm47blog} 
        & 28.78 & 31.07 \\
        GPT-OSS-120B~\citep{agarwal2025gpt} 
        & 14.57 & 30.47 \\
        DeepSeek-V4-Flash~\citep{deepseekai2026deepseekv4highlyefficientmilliontoken} 
        & 35.56 & 31.49 \\
        WebSailor-V2~\citep{li2025websailorv2bridgingchasmproprietary} 
        & 20.21 & 35.54 \\
        S1-Deep-Research~\citep{dong2026s1deepresearchsearchrealworldlonghorizon} 
        & 22.83 & 38.54 \\
        \bottomrule
        \end{tabular}
    
    \raggedright
    \rule{0pt}{3ex}\footnotesize{\textit{Note:} Acc. is final-answer accuracy, and Avg. Turns is the average number of tool-interaction turns per query. All models are evaluated under max\_turn = $60$ and context length = $131,072$.}
\end{table}


\begin{figure*}[!tp]
\centering
\scalebox{0.9}
{\includegraphics[width=\linewidth]{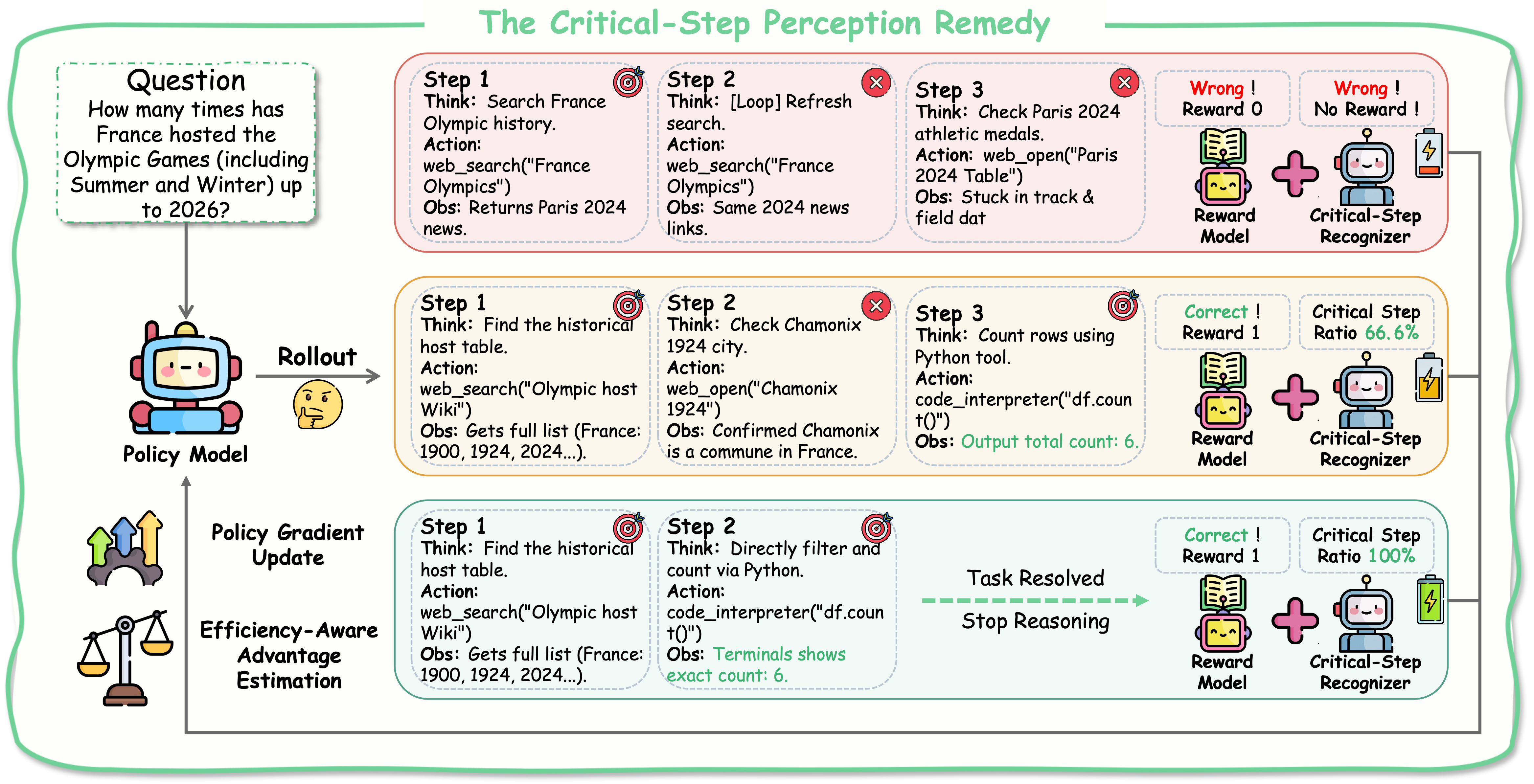}}
\caption{\textbf{Overview of CRISP}. CRISP introduces a critical-step recognizer identifying critical steps in policy rollouts. The recognized critical-step structure is combined with the outcome reward to perform efficiency-aware policy optimization, encouraging the policy to preserve necessary evidence while reducing redundant search behavior.}
\label{fig:main}
\end{figure*}

Recent work improves search efficiency along two lines. One regulates how often the agent acts, rewarding correct answers with fewer or more selective tool calls~\citep{wang2025acting, qian-etal-2025-smart, lin2025adasearch}; but penalizing call frequency treats all interactions alike, suppressing evidence-gathering steps along with redundant ones. A second selects which steps matter for more efficient tuning~\citep{chen-etal-2025-atlas}, yet defines criticality by structural or planning roles rather than evidential value. Complex search tasks call for a more evidence-centered notion: evidence-critical steps, defined as tool-interaction steps that provide or preserve evidence needed to support the final answer.

To address this gap, we propose \textbf{CRISP} (\textbf{Cri}tical \textbf{S}tep \textbf{P}erception), a framework for training deep search agents to identify evidence-critical steps. Figure~\ref{fig:main} illustrates the overall workflow. The key idea is to separate steps that provide or preserve evidence for the final answer from redundant tool interactions.
For example, in Figure~\ref{fig:main}, the first rollout gives a wrong answer and receives no efficiency bonus. The second rollout reaches the correct answer, but Step 2 performs an irrelevant intermediate check, so only two of its three tool interactions are recognized as critical, yielding a critical-step ratio of $2/3=66.6\%$. In contrast, every tool interaction in the third rollout obtains evidence for the final answer and is recognized as critical. In this way, CRISP evaluates not only final correctness, but also how much of a successful search process contributes useful evidence. To operationalize this idea, CRISP trains a critical-step recognizer to analyze full trajectories and predict critical tool-interaction steps. The recognizer is trained on search trajectories annotated by Backward Evidence Induction, in which a strong model traces each trajectory backward from the final answer, judging one candidate step at a time with all subsequently confirmed critical steps as context, and produces step-wise criticality labels and rationales. During policy optimization, CRISP applies the recognizer to successful rollouts, computing an efficiency-aware reward from the predicted numbers of critical steps and redundant steps. This reward encourages the policy to preserve evidence-critical interactions while reducing redundant search, rather than blindly shortening all trajectories. 
Empirically, on BrowseComp and HLE-Verified, CRISP maintains competitive final-answer accuracy while reducing average interaction turns by 15.1\% and 33.2\%, respectively. Our main contributions are summarized as follows:
\begin{itemize}
\item We introduce critical step perception for deep search agents, aiming to identify evidence-critical steps that form a compact evidence chain supporting the final answer.

\item We propose CRISP, an end-to-end framework that learns a critical-step recognizer and leverages its perception of critical steps to guide efficiency-aware policy optimization.

\item We empirically show that CRISP improves deep-search efficiency on challenging benchmarks, maintaining comparable answer accuracy while reducing interaction turns.
\end{itemize}

\begin{figure*}[!tp]
\centering
\scalebox{0.95}
{\includegraphics[width=\linewidth]{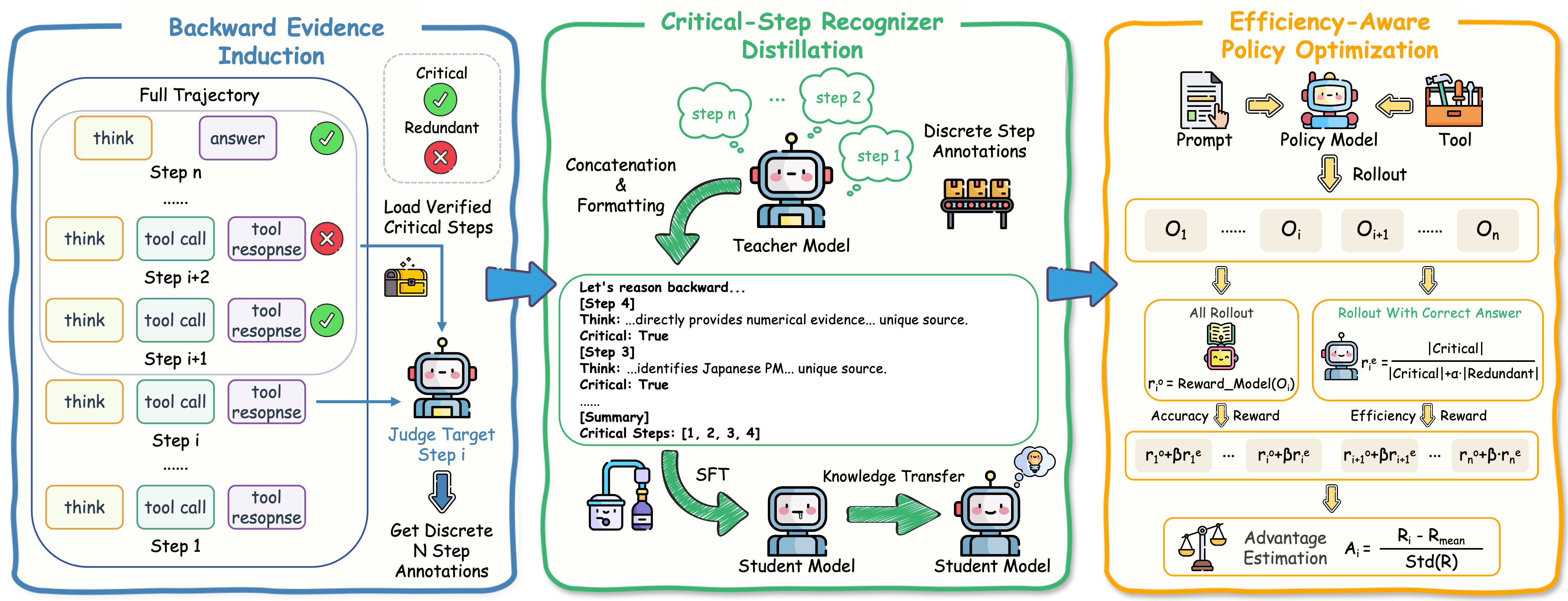}}
\caption{Detailed procedure of \textbf{CRISP}. 
\textbf{Left}: \textit{Backward Evidence Induction} obtains step-wise criticality annotations by traversing trajectories backward from the final response. 
\textbf{Middle}: \textit{Critical-Step Recognizer Distillation} formats these annotations into full-trajectory supervision and trains a student recognizer for single-pass critical-step analysis. 
\textbf{Right}: \textit{Efficiency-Aware Policy Optimization} uses the recognizer to estimate critical steps in successful rollouts and construct an efficiency-aware reward.}
\label{fig:method}
\end{figure*}

\section{Related Work}
\paragraph{Deep Search Agents}  These agents tackle knowledge-intensive questions through multi-turn web search, document retrieval, and information extraction, spending substantial test-time compute on many tool-call turns rather than parametric memory \citep{Guo_2025,zhu2025scalingtesttimecomputellm,huang2025deep}. A line of work elicits this behavior end-to-end via reinforcement learning rather than hand-crafted pipelines: Search-R1 \citep{jin2025searchr} and ReSearch \citep{chen2025researchlearningreasonsearch} learn to query from the final-answer signal alone, R1-Searcher \citep{song2025r1searcherincentivizingsearchcapability} uses a two-stage outcome-only scheme, and ZeroSearch \citep{sun2026zerosearchincentivizesearchcapability} replaces the live engine with an LLM-simulated retriever to cut cost and noise. The loop has since scaled from single lookups to long-horizon research and synthesis \citep{dong2026s1deepresearchsearchrealworldlonghorizon}. DeepResearcher \citep{zheng2025deepresearcherscalingdeepresearch} trains in real web environments, WebThinker \citep{li2025webthinkerempoweringlargereasoning} interleaves search, navigation, and drafting, SFR-DeepResearch \citep{nguyen2025sfrdeepresearcheffectivereinforcementlearning} trains a single autonomous agent without a fixed workflow, and Tongyi DeepResearch \citep{tongyideepresearchteam2026tongyideepresearchtechnicalreport} reports a full-stack deep-research system at scale. As the interaction horizon grows, the accumulated context becomes a bottleneck, and recent systems address it from the infrastructure side: Flash-Searcher \citep{qin2025flashsearcherfasteffectiveweb} restructures execution as a DAG for parallel speedup, ReSum \citep{wu2026resumunlockinglonghorizonsearch} periodically compresses history, and DeepMiner \citep{tang2025turnlimitstrainingdeep} holds a fixed window by discarding bulky tool logs while preserving the reasoning trace. Across these advances, however, stronger capability generally comes at the price of longer trajectories and more tool-call turns, so the interaction cost of deep search agents keeps growing, which is the efficiency bottleneck this work aims to address.

\paragraph{Efficiency Optimization for Deep Search}  Efficiency methods for deep search mainly differ in how they judge step usefulness. One group regulates how often the agent acts: OTC \citep{wang2025acting} rewards reaching the answer in fewer tool calls, while IKEA \citep{huang2025reinforcedinternalexternalknowledgesynergistic} retrieves externally only when internal knowledge is insufficient. A second reshapes the behavior distribution over a trajectory: ET-Agent \citep{chen2026etagentincentivizingeffectivetoolintegrated} corrects redundant tool use through a self-evolving data flywheel, Tool-Star \citep{dong2025toolstarempoweringllmbrainedmultitool} coordinates multiple tools with hierarchical rewards, and HDPO \citep{yan2026actwiselycultivatingmetacognitive} enforces economy only within already-correct trajectories. A third assigns step-level credit: DeepSearch \citep{wu2026deepsearchovercomebottleneckreinforcement} embeds Monte Carlo Tree Search into the RL loop for fine-grained credit assignment, ATLAS \citep{chen-etal-2025-atlas} trains only on the expert steps it deems critical, CSO \citep{li2026verifiedcriticalstepoptimization} marks a step critical when an alternative action verifiably flips failure into success, and MatchTIR \citep{qu2026matchtirfinegrainedsupervisiontoolintegrated} derives turn-level rewards by matching predicted against ground-truth traces. Closer to the tool calling, process reward models such as DataPRM \citep{qiu2026rewardingscientificprocessprocesslevel} and ToolPRMBench \citep{li2026toolprmbenchevaluatingadvancingprocess} push supervision from outcomes to individual tool-call steps. These works judge steps by length cost, outcome and structural role, or execution success. Our work differs in the criterion itself: while these criteria capture different aspects of step usefulness, none explicitly characterizes the evidential role of each interaction. 
We instead judge each step by whether the evidence it observes causally contributes to the final answer, letting CRISP reduce redundancy while preserving the steps that carry necessary evidence.

\section{Methodology}

We propose CRISP, a framework that locates evidence-critical steps in deep search trajectories and turns this perception into a training signal for efficiency. It has three stages: Backward Evidence Induction uses a strong model as a teacher to annotate which steps carry evidence for the final answer; Critical-Step Recognizer Distillation distills these annotations into a smaller student recognizer that labels critical steps for a full trajectory in a single pass; and Efficiency-Aware Policy Optimization applies the recognizer to successful rollouts and rewards trajectories with a higher proportion of critical steps. Figure~\ref{fig:method} illustrates the three stages, and all prompts are provided in Appendix~A.1.

Formally, a deep search agent answers a question $x$ with a trajectory $\tau=(s_1,\ldots,s_n)$ and a final answer $\hat{y}$, where each step $s_i=(t_i,a_i,o_i)$ collects the model's thought, action, and the observation returned by the environment. The last step $s_n$ carries the final answer $\hat{y}$ and invokes no tool. Based on this formulation, we address critical-step perception in the three subsections below.

\subsection{Backward Evidence Induction}
A straightforward way to obtain critical-step supervision is to ask a strong model to read the full trajectory and 
extract all evidence-critical steps. However, deep search trajectories can be long and noisy, containing repeated queries, weak clues, and observations that are never used in the final answer. Direct full-trajectory extraction therefore places a high demand on the teacher model's long-context understanding and structured extraction ability, making the annotation unstable. To obtain more reliable supervision, we introduce Backward Evidence Induction, which decomposes critical-step annotation into a sequence of local backward decisions.

The key idea is to judge each step with respect to the evidence that has already been confirmed as useful in the future steps. This motivates a backward pass: the teacher traverses the trajectory from $s_{n-1}$ to $s_1$, so that when a step is evaluated, all later steps have already been judged. For each step $i=n-1,\ldots,1$, the teacher model $T_\phi$ evaluates whether the current step is critical conditioned on the question, the final answer, the current step, the thought of the next step, and the already confirmed valid future steps:
\begin{equation}
(c_i,r_i)=T_\phi(x,\hat{y},s_i,t_{i+1},\mathcal{V}_{i+1})
\end{equation}
where $c_i\in\{0,1\}$ is the criticality label and $r_i$ is the teacher's brief rationale. Here $\mathcal{V}_{i+1}$ denotes steps already confirmed as evidence-critical among the future steps; it is initialized as $\mathcal{V}_n=\{s_n\}$, since the final-response step $s_n$ contains the answer $\hat{y}$. The next-step thought $t_{i+1}$ is used only as an auxiliary hint for what information from the current observation was noticed or used by the confirmed valid future steps.

A step is labeled as critical if it provides or preserves evidence for the final answer or directly enables a later confirmed step to access such evidence.
We further require this contribution to be marginal: if the same evidence is already provided more directly or completely by steps in $\mathcal{V}_{i+1}$, the current step is treated as non-critical.

After each judgment, we obtain $\mathcal{V}_i$ from $\mathcal{V}_{i+1}$ by adding the current step only when it is labeled critical: $\mathcal{V}_i=\mathcal{V}_{i+1}\cup\{s_i\}$ if $c_i=1$, and $\mathcal{V}_i=\mathcal{V}_{i+1}$ otherwise. The final set $\mathcal{V}_1$ therefore contains the trajectory steps that the teacher regards as evidence-critical.

This backward formulation makes teacher annotation more tractable. 
At each step, the teacher only needs to decide whether the current step provides additional evidence value beyond the confirmed valid future steps. 
The confirmed future steps serve as an explicit reference set, allowing the teacher to focus on local evidence contribution and redundancy rather than reconstructing the entire evidence chain from scratch. 
This step-wise procedure improves the stability and accuracy of critical-step supervision on long and noisy trajectories. 
\subsection{Critical-Step Recognizer Distillation}

Although Backward Evidence Induction provides reliable step-level annotations, it requires one teacher call per trajectory step, which becomes prohibitive for large-scale rollout analysis during policy optimization. To make critical-step perception efficient, we distill the teacher's backward annotation behavior into a smaller generative recognizer that reproduces the full backward analysis in a single pass.

After the backward traversal, the teacher has assigned each step $s_i$ a criticality label $c_i$ with a brief rationale $r_i$. We aggregate these local judgments into one structured target that mirrors the teacher's backward reasoning order:
\begin{equation}
\tilde{z}=\textsc{Format}\bigl((s_{n-1},c_{n-1},r_{n-1}),\ldots,(s_1,c_1,r_1)\bigr),
\end{equation}
where $\textsc{Format}$ serializes the per-step decisions from the last tool-interaction step to the first and appends a summary listing all critical steps. This target preserves the teacher's backward evidence induction while compressing $n-1$ teacher calls into a single generative response.

We train the student recognizer with supervised fine-tuning, taking the question $x$, the final answer $\hat{y}$, and the full trajectory $\tau$ as input and the formatted target $\tilde{z}$ as output. At inference time, the recognizer reads a trajectory once and the predicted critical steps are parsed from the final summary. This distillation amortizes the cost of Backward Evidence Induction and yields a reusable recognizer for trajectories beyond the offline annotation set. In the next section, we use it to estimate the numbers of critical and redundant steps in successful rollouts and construct an efficiency-aware reward for policy optimization.

\subsection{Efficiency-Aware Policy Optimization}

The distilled recognizer enables CRISP to use critical-step perception as a training signal for search agents. Standard outcome-based reinforcement learning only evaluates whether the final answer is correct, but does not distinguish between 
trajectories that gather the necessary evidence efficiently and trajectories that reach the same answer after many redundant tool interactions. To encourage more effective search behavior, we augment the answer correctness reward with an efficiency-aware reward computed from the recognizer's predicted critical steps.

For each question $x$, the current policy samples a group of rollouts:
\begin{equation}
\{(\tau_i,\hat{y}_i)\}_{i=1}^{G} \sim \pi_\theta(\cdot \mid x)
\end{equation}
where $G$ is the rollout group size. Let $R_i^{\mathrm{ans}}\in\{0,1\}$ denote the answer correctness reward. We apply the critical-step recognizer only to successful rollouts, i.e., rollouts with $R_i^{\mathrm{ans}}=1$. For a successful rollout $\tau_i=(s_{i,1},\ldots,s_{i,n_i})$, the recognizer predicts a critical-step indicator for each tool-interaction step:
\begin{equation}
\hat{c}_{i,j}\in\{0,1\}, \quad j=1,\ldots,n_i-1
\end{equation}
where the final response step is excluded from the efficiency reward.

We then define the number of predicted critical and redundant interaction steps as
\begin{equation}
K_i=\sum_{j=1}^{n_i-1}\hat{c}_{i,j},\qquad
T_i=(n_i-1)-K_i
\end{equation}
Here, $K_i$ measures how many tool-interaction steps are judged to contribute to the evidence chain, while $T_i$ measures the remaining redundant interaction steps. Based on these quantities, we define the efficiency-aware reward as
\begin{equation}
R_i^{\mathrm{crit}}
=
\frac{K_i}{K_i+\alpha T_i}
\end{equation}
where $\alpha$ controls the penalty strength for redundant steps. This reward favors trajectories in which a larger proportion of tool interactions are critical, while still allowing longer trajectories when additional steps contribute useful evidence.
\begin{table*}[t]
\centering
\small
\caption{
Main results on BrowseComp and HLE-Verified.
Official Acc. denotes reported model scores, and Official Setting lists the corresponding context length and maximum number of interaction turns.
Because official evaluations are reported only on the original HLE benchmark, the HLE-Verified Official Acc. column lists the corresponding HLE scores rather than HLE-Verified results.
All other results use our unified evaluation setting.
Down-arrow annotations denote relative reductions in average turns over Vanilla RL.
The asterisk indicates evaluation without tools, and a dash denotes an unreported or inapplicable result.
}
\label{tab:main_results}

\setlength{\tabcolsep}{5pt}

    \begin{tabular}{l cc ccc ccc}
    \toprule
    \multirow{2}{*}{\textbf{Model}}
    & \multicolumn{2}{c}{\textbf{Official Setting}}
    & \multicolumn{3}{c}{\textbf{BrowseComp}}
    & \multicolumn{3}{c}{\textbf{HLE-Verified}} \\
    \cmidrule(lr){2-3}
    \cmidrule(lr){4-6}
    \cmidrule(lr){7-9}
    & Context Length
    & Max Turns
    & Official Acc.$\uparrow$
    & Acc.$\uparrow$
    & Avg Turns$\downarrow$
    & Official Acc.$\uparrow$
    & Acc.$\uparrow$
    & Avg Turns$\downarrow$ \\
    \midrule

    GLM-4.5-Air
    & 128k
    & --
    & 21.30
    & 14.49
    & 31.23\phspace{21.8\%}
    & 10.60\textsuperscript{*}
    & 18.01
    & 8.34\phspace{21.8\%} \\

    GLM-4.7
    & 128k
    & --
    & 52.00
    & 28.78
    & 31.07\phspace{21.8\%}
    & 42.80
    & 33.56
    & 14.00\phspace{21.8\%} \\

    GPT-OSS-120B
    & --
    & --
    & --
    & 14.57
    & 30.47\phspace{21.8\%}
    & 19.00
    & 25.29
    & 4.24\phspace{21.8\%} \\

    DeepSeek-V4-Flash
    & 512k
    & 500
    & 53.50
    & 35.56
    & 31.49\phspace{21.8\%}
    & 40.30
    & 47.28
    & 16.77\phspace{21.8\%} \\

    WebSailor-V2
    & 128k
    & 100
    & 35.30
    & 20.21
    & 35.54\phspace{21.8\%}
    & 30.60
    & 35.40
    & 13.51\phspace{21.8\%} \\

    S1-Deep-Research
    & 128k
    & 150
    & 36.70
    & 22.83
    & 38.54\phspace{21.8\%}
    & 30.30
    & 27.82
    & 12.00\phspace{21.8\%} \\

    \midrule

    GLM-4.5-Air-Midtrain
    & 128k
    & 60/100
    & 17.59
    & 17.59
    & 11.53\phspace{21.8\%}
    & 17.09
    & 17.09
    & 4.38\phspace{21.8\%} \\

    \quad \textit{Vanilla RL}
    & 128k
    & 60/100
    & 36.22
    & 36.22
    & 30.95\phspace{21.8\%}
    & 33.02
    & 33.02
    & 9.86\phspace{21.8\%} \\

    \quad \textit{OTC-GRPO}
    & 128k
    & 60/100
    & 29.40
    & 29.40
    & 21.15\phspace{21.8\%}
    & 31.87
    & 31.87
    & 4.55\phspace{21.8\%} \\

    \quad \textit{\textbf{CRISP}}
    & 128k
    & 60/100
    & 35.69
    & 35.69
    & 26.28 \down{15.1\%}
    & 33.41
    & 33.41
    & 6.58 \down{33.2\%} \\

    \bottomrule
    \small
    \end{tabular}

\end{table*}
The final reward combines answer correctness with the efficiency-aware reward:
\begin{equation}
R_i = R_i^{\mathrm{ans}} + \lambda R_i^{\mathrm{ans}} R_i^{\mathrm{crit}}
\end{equation}
Since $R_i^{\mathrm{ans}}$ gates the efficiency-aware reward, unsuccessful rollouts receive no efficiency bonus. This prevents the policy from being rewarded for short but incorrect trajectories. In other words, CRISP does not reward acting less in isolation; it rewards successful trajectories whose tool interactions contain a higher proportion of evidence-critical steps.

We optimize the policy with GRPO \citep{shao2024deepseekmathpushinglimitsmathematical} using the final reward $R_i$. By incorporating critical-step perception into the reward, 
CRISP guides the policy toward reducing redundant search while preserving interactions needed to gather evidence for correct answers.

\section{Experiments}
\subsection{Experimental Setup}
\paragraph{Datasets and Models} 
Our trained policies are initialized from GLM-4.5-Air-Midtrain, which we obtain by mid-training GLM-4.5-Air \citep{5team2025glm45agenticreasoningcoding} for agentic deep search. We use Qwen3.5-397B-A17B~\citep{qwen3.5} as the teacher for Backward Evidence Induction and train a Qwen3.5-9B~\citep{qwen3.5} student recognizer with teacher-labeled data built from collected search trajectories and a subset of BrowseComp-Plus (BCP) \citep{chen2025browsecompplusfairtransparentevaluation} trajectories. We evaluate CRISP on two deep-search benchmarks that require multi-step evidence gathering rather than single-shot retrieval: BrowseComp \citep{wei2025browsecomp}, which targets hard-to-find web evidence and is closer to the BCP trajectories used to train the recognizer, and the pure-text subset of HLE-Verified \citep{zhai2026hleverifiedsystematicverificationstructured}, a reliability-enhanced version of Humanity's Last Exam (HLE)~\citep{phan2025lastexam} that contains challenging multi-domain QA-style problems and serves as a cross-domain test of whether the learned critical-step structure generalizes beyond BCP-specific patterns. The resulting datasets contain approximately $1.3$k questions for BrowseComp and $2.2$k questions for HLE-Verified. For each benchmark, we randomly partition the questions into an $80\%$ training split and a $20\%$ held-out test split. Agents operate in a tool-augmented environment: all tasks allow external web search and browsing, while HLE-Verified additionally permits executable code for computation-oriented reasoning.

\paragraph{Baselines} 
We compare CRISP with these paradigms:
\begin{itemize}
\item \textbf{Foundation Models with Tools} We test general-purpose LLMs under our agent scaffold, including GPT-OSS-120B \citep{agarwal2025gpt}, GLM-4.5-Air \citep{5team2025glm45agenticreasoningcoding}, GLM-4.7 \citep{glm47blog}, and DeepSeek-V4-Flash \citep{deepseekai2026deepseekv4highlyefficientmilliontoken}. 
\item \textbf{Open-Source Deep-Search Agents} We compare with recent open-source deep search agents trained for long-horizon search, including S1-Deep-Research \citep{dong2026s1deepresearchsearchrealworldlonghorizon} and WebSailor-V2 \citep{li2025websailorv2bridgingchasmproprietary}.
\item \textbf{Reward-Design Baselines} We train policies from the same GLM-4.5-Air-Midtrain backbone under our RL setup, but vary the reward design. \textit{Vanilla RL} uses only the outcome-based GRPO reward \citep{shao2024deepseekmathpushinglimitsmathematical}, and \textit{OTC-GRPO} \citep{wang2025acting} is an efficiency-oriented variant that penalizes tool-use cost to reduce interactions.
\end{itemize}

\paragraph{Implementation Details}
We train the critical-step recognizer with full-parameter supervised fine-tuning and apply it to each successful rollout during policy optimization. Appendix~A.2 provides a standalone evaluation of the recognizer and its critical-step extraction quality. For the efficiency-aware reward, we set $\alpha=0.7$ to favor trajectories with a higher proportion of evidence-critical interactions, discouraging redundant exploration without uniformly penalizing tool use; a sensitivity analysis of $\alpha$ is provided in Appendix~A.3. We set a small auxiliary reward weight $\lambda=0.1$, so that the efficiency signal guides successful trajectories without overriding answer correctness. We optimize the policy with GRPO. During training, all methods use a $131{,}072$-token context, temperature $1.0$, and top-$p$ $1.0$. Official evaluations often allow large interaction budgets to probe model capability, but practical deployments must also account for user-facing latency and the cost of search calls and inference. We therefore use tighter budgets of $60$ turns for BrowseComp and $100$ turns for HLE-Verified. Additional training configurations are provided in Appendix~A.4.

\paragraph{Evaluation} Evaluation uses the same experimental configuration as training. For metrics, we report \textbf{Acc}, the final-answer accuracy scored by task-specific LLM judges (GLM-4.7 for BrowseComp and GPT-OSS-120B for HLE-Verified), and \textbf{Avg Turns}, the average tool-interaction turns per question. These metrics measure task effectiveness and interaction efficiency, respectively. All results are averaged over three independent runs.

\subsection{Main Results}
\paragraph{Overall Performance}
Table~\ref{tab:main_results} presents the main results on BrowseComp and HLE-Verified. Starting from the same backbone, both Vanilla RL and CRISP substantially improve answer accuracy, confirming the effectiveness of reinforcement learning for deep search. More importantly, CRISP consistently improves interaction efficiency while preserving answer quality: accuracy remains comparable on BrowseComp and slightly higher on HLE-Verified, while the average number of turns drops by 15.1\% and 33.2\%, respectively. These results show that CRISP reduces redundant interactions without suppressing the interactions needed to gather evidence for correct answers. The open-source models provide additional reference points under the same agent scaffold, showing substantial variation in both answer accuracy and interaction cost.

\paragraph{Efficiency Trade-off}
OTC-GRPO compresses trajectories more aggressively, but loses answer accuracy on both benchmarks. Compared with it, CRISP keeps slightly longer trajectories, with 5.13 and 2.03 more turns on BrowseComp and HLE-Verified, but gains 6.29 and 1.54 accuracy points, respectively. It suggests that the preserved interactions of CRISP often carry necessary evidence rather than redundancy. The accuracy degradation of OTC-GRPO is also more pronounced on BrowseComp, where the trajectories are substantially longer, indicating that aggressive compression can be particularly harmful in longer-horizon search. Overall, uniformly favoring fewer tool calls may suppress not only redundant interactions but also evidence-critical steps necessary for solving difficult questions. By distinguishing critical from redundant interactions, CRISP reaches a better accuracy-efficiency trade-off.

\subsection{Analysis}

\paragraph{Functional Validation of Critical-Step Recognition}
We further use trajectory intervention to examine whether the critical steps identified by the recognizer are more important for recovering the final answer. To avoid policy-recognizer co-adaptation, we use trajectories from Vanilla RL rather than CRISP. We randomly sample 100 examples from the HLE-Verified test set. For each example, we remove the original final-response step to prevent answer leakage. We then let the same policy continue generation with tool access under three contexts: the full trajectory prefix, only the predicted critical steps, or only the predicted redundant steps. The full-context setting controls for the variance introduced by regeneration itself, while the key comparison is between the critical-only and redundant-only settings. We report final-answer accuracy and the average number of total interaction turns of the resulting trajectories.

As shown in Table~\ref{tab:critical_step_intervention}, the full context achieves accuracy and average turns close to the original rollout, indicating limited degradation from regeneration. Retaining only the predicted critical steps preserves accuracy close to the full context while completing trajectories with markedly fewer turns, whereas retaining only the predicted redundant steps yields the lowest accuracy despite consuming the most turns. This asymmetric effect shows that the recognizer's critical-step partition has functional meaning: the predicted critical steps carry information more relevant to the final answer, whereas the redundant steps lack this information, forcing the policy to spend additional turns searching to recover it.

\begin{table}[t]
\centering
\small
\caption{
Trajectory intervention results. $\Delta$Acc. and $\Delta$Turns denote
differences relative to the Full Context setting.
}
\label{tab:critical_step_intervention}
\setlength{\tabcolsep}{3pt}
\renewcommand{\arraystretch}{1.08}

\begin{tabular}{lcccc}
\toprule
Context
& Acc. (\%)
& Avg. Turns
& $\Delta$Acc.
& $\Delta$Turns \\
\midrule
Original Rollout & 36.00 & 10.53 & $+1.00$ & $-0.56$ \\
Full Context     & 35.00 & 11.09 & --      & --      \\
Critical Only    & 33.00 & 8.14  & $-2.00$ & $-2.95$ \\
Redundant Only   & 27.00 & 11.67 & $-8.00$ & $+0.58$ \\
\bottomrule
\end{tabular}

\end{table}

\paragraph{Evidence Density}
Since critical steps are defined with respect to the evidence supporting the final answer, we analyze correctly answered evaluation trajectories and compute the proportion of tool-interaction steps predicted as critical by the recognizer. Compared with Vanilla RL, CRISP increases the average critical-step ratio from 34.06\% to 41.09\% on BrowseComp and from 42.06\% to 53.09\% on HLE-Verified, corresponding to relative improvements of 20.6\% and 26.2\%, respectively. 
These consistent gains indicate that CRISP produces trajectories with higher evidence density. Together with the reductions in average turns reported in Table~\ref{tab:main_results}, they show that CRISP improves efficiency by selectively reducing redundant interactions while preserving critical steps, rather than by indiscriminately shortening successful search trajectories.

\paragraph{Tool-Use Behavior}
To examine whether CRISP merely suppresses tool use or instead reshapes the interaction pattern, we report the average number of calls by tool type in Table~\ref{tab:tool_behavior}. On BrowseComp, CRISP reduces search calls while increasing browsing calls, suggesting a shift from repeated broad querying toward more focused access to retrieved documents. This increase in browsing calls is particularly informative: CRISP does not simply learn to call tools less often, but reallocates interactions toward tools that are more likely to provide evidence once candidate documents have been retrieved. On HLE-Verified, CRISP reduces search, browsing, and code calls simultaneously, indicating that it removes unnecessary interactions more broadly across both information seeking and computation. 
This opposite trend reflects the different search demands of the two benchmarks. BrowseComp often requires long-horizon web exploration over diverse and noisy evidence sources, so after reducing repeated search calls, CRISP allocates more interactions to browsing candidate documents and extracting useful evidence. In contrast, HLE-Verified involves shorter trajectories with more concentrated information needs, where CRISP can reduce redundant follow-up browsing together with search and code calls. Overall, these benchmark-specific changes suggest that CRISP reshapes the model's tool-use pattern, preserving interactions that are likely to support evidence acquisition while reducing those that mainly contribute redundant exploration.

\begin{table}[t]
\centering
\small
\caption{Average number of tool calls per question by tool type. $\Delta$ denotes the change from Vanilla RL to CRISP.}
\setlength{\tabcolsep}{3pt}
\begin{tabular}{llrrr}
\toprule
Benchmark & Tool & Vanilla RL & CRISP & $\Delta$ \\
\midrule
BrowseComp & Search   & 28.98 & 23.46 & -5.52 \\
BrowseComp & Browsing & 1.15  & 1.95  & +0.80 \\
\midrule
HLE-Verified & Search   & 4.01 & 2.64 & -1.37 \\
HLE-Verified & Browsing & 1.48 & 1.13 & -0.35 \\
HLE-Verified & Code     & 3.40 & 1.82 & -1.58 \\
\bottomrule
\end{tabular}

\label{tab:tool_behavior}
\end{table}

\paragraph{Exploration after Reward Removal}
A potential concern is that the efficiency-aware reward may reduce interaction cost by suppressing policy exploration. To examine this possibility, we remove the efficiency-aware reward from CRISP and continue training both the Vanilla RL and CRISP policies on HLE-Verified under the same outcome-only objective. After reward removal, the CRISP policy retains a substantially higher policy entropy of 0.672, compared with 0.405 for Vanilla RL. Higher entropy indicates that CRISP maintains a broader action distribution rather than converging to a narrowly deterministic strategy. At the same time, CRISP achieves an answer accuracy of 33.71\%, which is comparable to the 33.56\% obtained by Vanilla RL. Its average number of interaction turns also remains markedly lower, decreasing from 10.18 for Vanilla RL to 6.43 for CRISP. The coexistence of higher entropy, comparable accuracy, and shorter trajectories suggests that CRISP does not obtain efficiency through low-entropy policy collapse. Instead, the learned policy preserves diverse action choices while becoming more selective about when additional interactions are useful. The persistence of this behavior after removing the efficiency-aware reward further demonstrates that CRISP induces a lasting change in the policy rather than a temporary response to the auxiliary reward signal.

\begin{figure}[!tp]
\centering
\scalebox{0.95}
{\includegraphics[width=\linewidth]{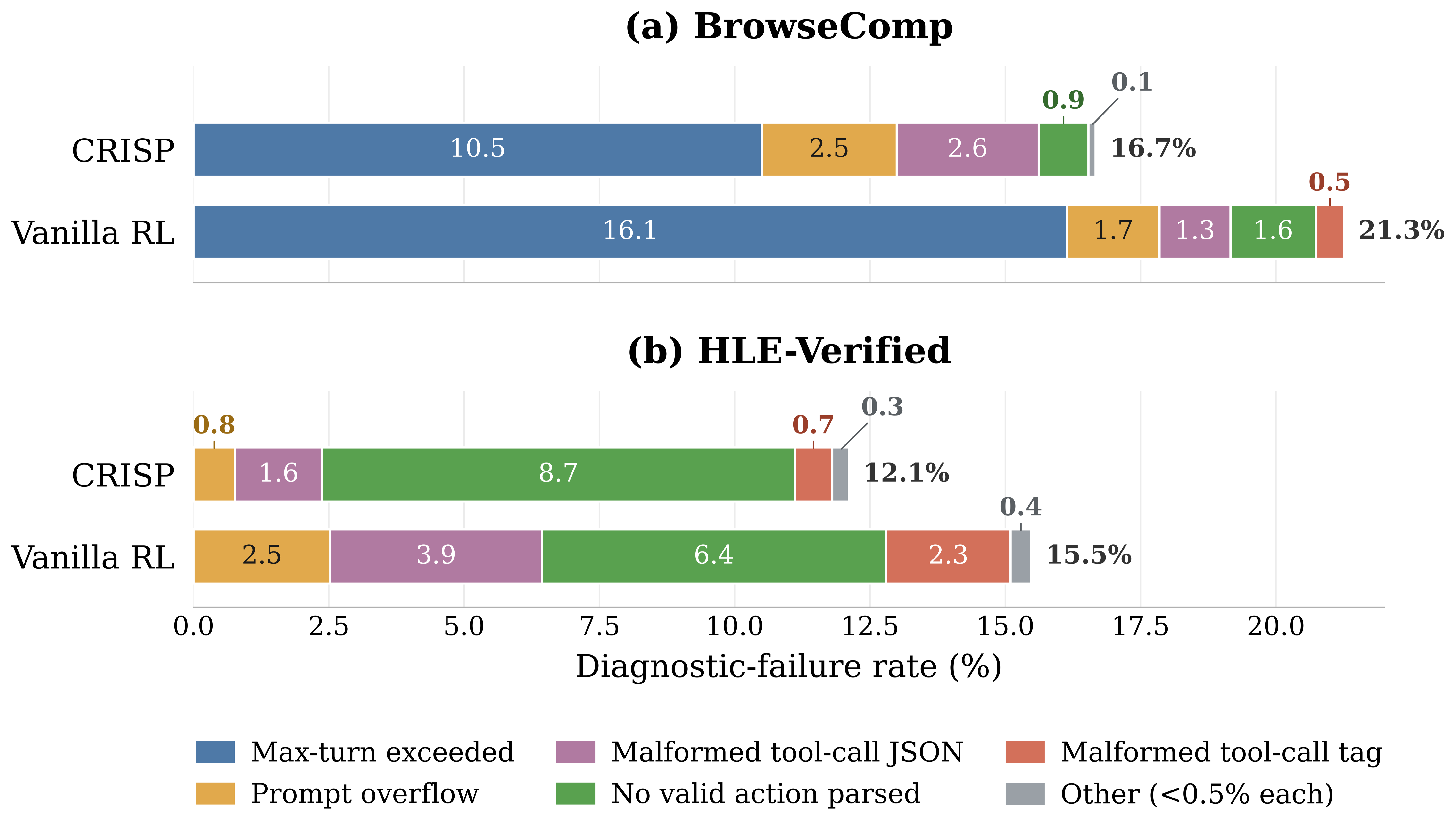}}
\caption{Composition of diagnostic failure modes on BrowseComp and HLE-Verified. Each stacked segment shows one failure cause, and the number at the right of each bar is the total failure rate; causes below 0.5\% are grouped as Other.}
\label{fig:failure_diagnostics}
\end{figure}

\paragraph{Failure Diagnostics}
To examine whether CRISP's efficiency gains also improve execution reliability, we compute the failure rate, defined as the percentage of evaluation queries where the agent produces no valid final response due to interrupted execution or unparsable actions; incorrect but valid final answers are not counted. As shown in Figure~\ref{fig:failure_diagnostics}, CRISP lowers the failure rate, with relative reductions of 21.6\% on BrowseComp and 21.8\% on HLE-Verified.

Figure~\ref{fig:failure_diagnostics} further decomposes these failed trajectories by diagnostic cause. \textit{Max-turn exceeded} means that the interaction budget is exhausted before a final response is produced, while \textit{prompt overflow} indicates that the accumulated trajectory exceeds the context window. \textit{Malformed tool-call JSON} and \textit{malformed tool-call tag} denote invalid tool-call content and invalid enclosing tag structure, respectively. \textit{No valid action parsed} refers to cases in which the model attempts no tool call at all and commits to no final answer.

This breakdown suggests that CRISP mainly reduces failures caused by excessive trajectory growth and redundant interactions. On BrowseComp, the dominant max-turn failure becomes less frequent, consistent with the reduction in repeated search calls observed above. 
On HLE-Verified, prompt overflow and tool-formatting failures both decline, indicating that shorter and more selective trajectories reduce context accumulation while also limiting the number of tool-call generation points where formatting errors can occur. Although no valid action parsed accounts for a substantial share of the remaining failures, CRISP still achieves a lower overall failure rate, indicating that its selective reduction of redundant interactions improves end-to-end reliability. Overall, by reducing redundant tool interactions, CRISP not only improves efficiency but also makes execution more reliable.

\section{Conclusion}
In this work, we present CRISP, a framework for training efficient deep search agents through critical-step perception. CRISP uses Backward Evidence Induction to annotate search trajectories by tracing backward from the final answer and judging whether each tool-interaction step provides critical evidence. It then distills these step-wise judgments into a generative recognizer that analyzes full trajectories and summarizes their critical steps. During policy optimization, the recognizer guides the agent toward successful trajectories with a higher proportion of evidence-critical interactions and fewer redundant tool calls. Experiments on BrowseComp and HLE-Verified show that CRISP maintains competitive accuracy while reducing average interaction turns by 15.1\% and 33.2\%, respectively. These results suggest that critical-step perception provides a practical path toward deep search agents with effective tool use and reduced redundancy.

\bibliography{arxiv}


\appendix
\onecolumn
\subsection{A.1 Prompts}

We provide the full prompt templates used in our pipeline. In
Stage~1 (Backward Evidence Induction), the \textbf{teacher critic}
(Prompt~1) is queried once per candidate step: starting from the final
answer and tracing the trajectory backward, it judges whether a single
\emph{current step} is evidence-critical, conditioned on the confirmed
valid future steps retained so far, and emits a compact JSON verdict.
Since per-step querying is accurate but costly, Stage~2 (Critical-Step
Recognizer Distillation) distills the teacher into a \textbf{student
recognizer} (Prompt~2) that ingests the entire trajectory at once and
reproduces the same backward judgement as a single chain of thought
terminating in a critical-step list. Both prompts serialize trajectory
content through a shared step schema (Figure~\ref{fig:step-format}):
the teacher's \texttt{\{Valid\_Future\_Steps\_String\}} and the
student's \texttt{\{Full\_Trajectory\}} are instantiated from the same
\texttt{[Step~N]}\,/\,\emph{Thought}\,/\,\emph{Action}\,/\,\emph{Observation}
template. Runtime placeholders (e.g., \texttt{\{Question\}},
\texttt{\{Final\_Answer\}}) are filled per instance. Here \texttt{\{Last\_Step\_Num\}} is the final-answer step index, where
backward evaluation begins.

\begin{promptbox}{Prompt 1 — Teacher Critic (per-step backward evaluation, Stage~1)}
\begin{Verbatim}[fontsize=\small, breaklines=true, breakanywhere=true, commandchars=\!\^\~]
You are an expert in AI trajectory analysis and reward modeling. Your task is to
evaluate whether the "Current Step" is a critical and non-redundant part of the
minimal evidence chain that preserves the necessary evidence for the final answer.

!textbf^[Input Information]~
1. Target Question: {Question}
2. Final Answer (Step {Last_Step_Num}): {Final_Answer}
3. Current Step to Evaluate (Step {Current_Step_Num}):
   Thought: {Current_Thought}
   Action: {Current_Action}
   Observation: {Current_Observation}
4. Thought From Next Step (Step {Next_Step_Num}):
   {Next_Thought}
5. Confirmed Valid Future Steps (already extracted critical steps, back to front):
{Valid_Future_Steps_String}   # see Figure 3 for the step format

!textbf^[Evaluation Criteria]~
To be marked as critical (`true`), the Current Step MUST satisfy BOTH
[Condition A] AND [Condition B].

!textbf^[Condition A]: Evidence Value~
The step must satisfy at least ONE of the following:
1. Direct Evidence Contribution:
   Its `Observation` directly contains facts, data, document content, or document
   IDs that are necessary for supporting the Final Answer.
2. Evidence-Enabling Contribution:
   Its `Observation` provides a document identifier, URL, entity anchor, or other
   specific pointer that directly enables later steps to access, open, or identify
   necessary evidence.
   Mere help for search planning, query writing, or reasoning is NOT sufficient.

When judging whether the Current Step provides useful information, you may use the
Thought From Next Step as an auxiliary hint for what information from the Current
Step was actually noticed and used downstream. However, the Thought From Next Step
is only supporting evidence and does NOT by itself make the Current Step critical.

!textbf^[Condition B]: Non-Redundant Marginal Evidence Gain~
Because we are evaluating backward, the Current Step is redundant if its necessary
evidence, or its evidence-enabling contribution, is already contained more directly
or more completely in the [Confirmed Valid Future Steps].
Use the following redundancy rules:
- If a later confirmed valid step already contains the same necessary evidence more
  directly or more completely, prefer the later step.
- If the Current Step only provides a weaker clue, but a later step provides the
  actual necessary evidence, mark the Current Step as false.
- If the later steps only partially overlap, or do not clearly replace the Current
  Step's evidence value, keep the Current Step.

!textbf^[Tool-Specific Guidance]~
- Content-access steps are stronger evidence carriers than broad search-result steps.
- If a content-access step contains necessary evidence, prefer keeping it unless
  later valid steps clearly contain the same evidence.
- A search-result step should be kept only if:
  (a) it already contains necessary evidence that is not clearly recovered later, OR
  (b) it uniquely and directly points to necessary evidence later valid steps rely on.
- Do NOT keep a search step merely because it suggested a direction, keyword, or
  candidate.

!textbf^[Strict Exclusions] (Must be `false`)~
- Mine-sweeping or negative results: steps that only rule out an incorrect option or
  confirm what is NOT the answer.
- Trial-and-error, API errors, 404 pages, or hallucinations.
- Steps that only help later Thought formulation, but do not carry or directly enable
  necessary evidence.

!textbf^[Output Format]~
Output your analysis strictly in the following JSON format:
{
  "brief_reasoning": "Briefly explain: 1) whether this step provides direct evidence
   or evidence-enabling value; 2) what necessary evidence it supports; 3) whether
   later valid steps clearly replace it.",
  "is_critical": true/false
}
\end{Verbatim}
\end{promptbox}
\captionof{figure}{Teacher critic prompt (Stage~1). The teacher evaluates one candidate
step at a time, conditioned on the confirmed valid future steps, and emits a JSON criticality
verdict.}
\label{fig:prompt-teacher}

\begin{promptbox}{Prompt 2 — Student Recognition Model (full-trajectory backward labeling, Stage~2)}
\begin{Verbatim}[fontsize=\small, breaklines=true, breakanywhere=true, commandchars=\!\^\~]
You are an expert in AI trajectory analysis and reward modeling. Your task is to
label which steps in a full agent trajectory are critical, non-redundant parts of
the minimal evidence chain that preserves the necessary evidence for the final answer.

!textbf^[Input Information]~
1. Target Question: {Question}
2. Final Answer (Step {Last_Step_Num}): {Final_Answer}
3. Full Trajectory: {Full_Trajectory}   # see Figure 3 for the step format

!textbf^[Goal]~
Identify the minimal set of trajectory steps that are necessary to preserve the
evidence used to support the Final Answer.

!textbf^[Evaluation Criteria]~
A step is `critical` if it satisfies BOTH [Condition A] and [Condition B].

!textbf^[Condition A]: Evidence Value~
The step must satisfy at least ONE of the following:
1. Direct Evidence Contribution:
   Its `Observation` directly contains facts, data, document content, or document
   IDs that are necessary for supporting the Final Answer.
2. Evidence-Enabling Contribution:
   Its `Observation` provides a document identifier, URL, entity anchor, or other
   specific pointer that directly enables later steps to access, open, or identify
   necessary evidence.
   Mere help for search planning, query writing, or reasoning is NOT sufficient.

Later steps may be used as auxiliary evidence to infer what information from an
earlier step was actually noticed and used downstream. However, being noticed or
used later does NOT by itself make an earlier step critical. An earlier step is
critical only if it is necessary for preserving the minimal evidence chain.

!textbf^[Condition B]: Non-Redundant Marginal Evidence Gain~
A step is redundant if its necessary evidence, or its evidence-enabling contribution,
is already provided more directly or more completely by other retained steps.
Use the following redundancy rules:
- If a later step already contains the same necessary evidence more directly or more
  completely, prefer the later step.
- If a step only provides a weaker clue, but another retained step provides the
  actual necessary evidence, mark the weaker step as non-critical.
- If later steps only partially overlap, or do not clearly replace the step's
  evidence value, keep the step.

!textbf^[Tool-Specific Guidance]~
- Content-access steps are stronger evidence carriers than broad search-result steps.
- If a content-access step contains necessary evidence, prefer keeping it unless
  another retained step clearly contains the same evidence.
- A search-result step should be kept only if:
  (a) it already contains necessary evidence that is not clearly recovered later, OR
  (b) it uniquely and directly points to a necessary evidence document that later
      retained steps rely on.
- Do NOT keep a search step merely because it suggested a direction, keyword, or
  candidate.

!textbf^[Strict Exclusions] (Must be `non_critical`)~
- Mine-sweeping or negative results: steps that only rule out an incorrect option or
  confirm what is NOT the answer.
- Trial-and-error, API errors, 404 pages, or hallucinations.
- Steps that only help later thought formulation, but do not carry or directly enable
  necessary evidence.

!textbf^[Output Format]~
Please output the analysis in a step-by-step thinking chain, from the last step to
the first step. For each step, include:
  [Step N]
  Thought: Describe your reasoning for whether this step is critical.
  Critical: True/False
After all steps, provide a [Step Summary] section listing all critical step numbers:
  [Step Summary]
  Critical Steps: [Critical_Steps_List]
\end{Verbatim}
\end{promptbox}
\captionof{figure}{Student recognizer prompt (Stage~2). The distilled recognizer reads the
entire trajectory at once and reproduces the backward judgement as a single chain of thought.}
\label{fig:prompt-student}

\begin{promptbox}{Shared Trajectory Step Format (used by both prompts)}
\begin{Verbatim}[fontsize=\small, breaklines=true, breakanywhere=true, commandchars=\!\^\~]
Both the teacher's {Valid_Future_Steps_String} and the student's {Full_Trajectory}
are serialized in the following unified step format (steps are listed back to front
for the teacher, and front to back for the student):

[Step N]
Thought: <the agent's reasoning at this step>
Action: <the tool call issued at this step>
Observation: <the raw content returned by the tool>
[Step N-1]
Thought: ...
Action: ...
Observation: ...
...
\end{Verbatim}
\end{promptbox}
\captionof{figure}{The shared trajectory step schema used by both prompts. The teacher's
confirmed valid future steps and the student's full trajectory are both serialized through
this template.}
\label{fig:step-format}

\subsection{A.2 Evaluation of the Critical-Step Recognizer}
\label{app:recognizer_validation}

In this section, we evaluate whether the critical-step recognizer can
perceive evidence-critical steps in deep search trajectories. The
evaluation is conducted on BrowseComp-Plus (BCP), which provides gold
evidence document IDs for each question. We treat a trajectory step as
evidence-critical if its observation contains a gold evidence document ID,
and use these gold-bearing steps as ground truth to assess the steps
extracted by each method. Since critical-step perception is defined with
respect to the evidence chain supporting the final answer, we only evaluate
trajectories with correct final answers.

\paragraph{Evaluation protocol} We evaluate under two settings. On the
\emph{full} BCP set, we analyze different critical-step extraction
strategies on correct GLM-4.7 trajectories, comparing annotation strategies
and extraction feasibility (Table~\ref{tab:annotation_strategy_analysis}).
This comparison does not involve the trained recognizer, so it can use all
correct trajectories. To evaluate the distilled recognizer itself, we use a
\emph{held-out} BCP split: because the teacher-labeled SFT data for the
recognizer are built from collected search trajectories together with a
subset of BCP trajectories, the recognizer must be tested on trajectories
that are not included in its SFT data to avoid leakage
(Table~\ref{tab:heldout_recognizer_eval}).

\paragraph{Evaluation metrics} We evaluate extracted critical steps by matching them against the gold evidence document IDs provided by BCP. Throughout this evaluation, we operationalize \emph{evidence} as the gold
evidence document IDs provided by BCP: a trajectory step is said to
\emph{bear evidence} if and only if its observation contains at least one
gold evidence document ID. All metrics below are therefore computed by matching document
IDs rather than by judging semantic content. Let $M$ denote the number of evaluated trajectories, indexed by $j$. Each trajectory answers a question of BCP. For each trajectory $j$, let $\mathcal{G}_j$ denote the gold evidence document IDs set of the question, and let $\mathcal{D}^{\mathrm{traj}}_j$ denote all document IDs that appear in the original full trajectory. Given an extraction method, we parse the predicted critical steps and collect the document IDs appearing in their observations, denoted as $\mathcal{D}^{\mathrm{ext}}_j$. If the method fails to produce a valid and parsable critical-step list, we set $\mathcal{D}^{\mathrm{ext}}_j=\emptyset$ for recall and coverage computation.

\textbf{Success rate} measures the percentage of trajectories for which a valid critical-step list can be successfully parsed. Failures may be caused by output truncation, invalid formatting, or the model being distracted by long noisy trajectories.

\textbf{Origin recall} measures the evidence coverage of the original trajectory before extraction:
\begin{equation}
\mathrm{OriginRecall}
=
\frac{1}{M}
\sum_{j=1}^{M}
\frac{
|\mathcal{D}^{\mathrm{traj}}_j \cap \mathcal{G}_j|
}{
|\mathcal{G}_j|
}.
\end{equation}
This metric reflects the upper bound imposed by whether the gold evidence documents appear in the trajectory.

\textbf{Extract recall} measures the fraction of gold evidence documents covered by the extracted critical steps:
\begin{equation}
\mathrm{ExtractRecall}
=
\frac{1}{M}
\sum_{j=1}^{M}
\frac{
|\mathcal{D}^{\mathrm{ext}}_j \cap \mathcal{G}_j|
}{
|\mathcal{G}_j|
}.
\end{equation}

\textbf{Evidence coverage accuracy} is a stricter case-level metric that measures whether all gold evidence documents are covered:
\begin{equation}
\mathrm{CoverageAcc}
=
\frac{1}{M}
\sum_{j=1}^{M}
\mathbb{I}
\left[
\mathcal{G}_j \subseteq \mathcal{D}^{\mathrm{ext}}_j
\right].
\end{equation}

\textbf{Step hit} measures what fraction of the steps our method extracts as
critical are genuinely critical, i.e., the precision of the perceived
critical steps. Since we operationalize a genuinely critical step as one
that bears evidence, we count an extracted step as a hit if its observation
contains at least one gold evidence document ID. Let $\mathcal{C}_j$ be the
set of extracted critical steps for trajectory $j$, and let $\mathcal{D}(s)$
denote the document IDs appearing in step $s$. We compute step hit as:
\begin{equation}
\mathrm{StepHit}
=
\frac{
\sum_{j=1}^{M}
\sum_{s\in \mathcal{C}_j}
\mathbb{I}
\left[
\mathcal{D}(s)\cap \mathcal{G}_j \neq \emptyset
\right]
}{
\sum_{j=1}^{M}
|\mathcal{C}_j|
}.
\end{equation}

Together, these metrics capture what a good critical-step recognizer needs:
completeness and precision. \emph{Origin recall} is the upper bound fixed by
the trajectory, so the completeness metrics, extract recall and coverage
accuracy (the latter being the stricter all-or-nothing version), are read
against it to show how much recoverable evidence is actually perceived.
\emph{Step hit} measures precision: what fraction of the extracted steps are
genuinely critical rather than redundant. A recognizer is effective when
recall and coverage stay close to the origin-recall ceiling at a high step
hit, perceiving the evidence chain completely without inflating it. Success
rate is a prerequisite, since an unparsable output perceives nothing.

\paragraph{Annotation strategy analysis} We first compare different
critical-step extraction strategies. \textbf{Teacher Direct} asks the teacher
model (Qwen3.5-397B-A17B) to read the full trajectory and extract all
critical steps in one pass. \textbf{Teacher Backward} applies the same teacher model with our proposed
Backward Evidence Induction (BEI, Stage~1), judging one candidate step at a
time from the final response backward. \textbf{Student Direct} uses the untrained student backbone (Qwen3.5-9B), the
same model later distilled into our recognizer, to perform direct
full-trajectory extraction. It serves as a before-distillation reference for
whether a smaller model can perform this task on its own.

\begin{table*}[t]
\centering
\small{
    \begin{tabular}{lccccc}
    \toprule
    Method & Success & Extract Recall & Coverage Acc. & Step Hit & Complexity \\
    \midrule
    Teacher Direct   & 100.0\% & 0.7780 & 72.94\% & 0.9712 & $O(1)$ \\
    Teacher Backward & 100.0\% & \textbf{0.7995} & \textbf{80.30\%} & \textbf{0.9803} & $O(n)$ \\
    Student Direct     & 38.7\%  & 0.2746 & 24.24\% & 0.9361 & $O(1)$ \\
    \bottomrule
    \end{tabular}
}

\caption{
Annotation strategy analysis on correct BCP trajectories generated by GLM-4.7.
The original full trajectories have an origin recall of 0.8422.
Backward annotation provides better evidence coverage than direct extraction
with the same teacher.
}
\label{tab:annotation_strategy_analysis}
\end{table*}

The results show that backward step-wise annotation provides more reliable
teacher supervision. Compared with direct extraction using the same teacher,
backward extraction improves both extract recall and evidence coverage
accuracy while maintaining a perfect extraction success rate. This suggests
that decomposing trajectory annotation into local backward decisions helps
the teacher perceive evidence-critical steps more accurately from long and
noisy trajectories.

The comparison with the untrained smaller model further motivates the need
for distillation. Although direct extraction only requires a single model
call, the untrained smaller model succeeds on only a minority of trajectories
and obtains substantially lower evidence coverage. This indicates that
critical-step extraction is not a trivial formatting task for smaller models,
especially under long full-trajectory inputs.

\paragraph{Held-out recognizer evaluation} We then evaluate the distilled
student recognizer on the held-out BCP split. The student recognizer performs
single-pass full-trajectory critical-step analysis, while the backward teacher uses step-wise annotation. This comparison tests whether the student
can absorb the teacher's backward evidence judgment behavior while avoiding
the $O(n)$ cost of teacher annotation. Note that this table and the
annotation strategy analysis in Table~\ref{tab:annotation_strategy_analysis}
are computed on different BCP subsets, so the teacher numbers are not directly
comparable across the two tables; each table should be read as an internally
consistent comparison.

\begin{table*}[t]
\centering
\scalebox{0.92}{
\begin{tabular}{lccc}
\toprule
Method & Extract Recall & Coverage Acc. & Step Hit \\
\midrule
Teacher Backward   & 0.7891 & 83.51\% & 0.9791 \\
Student Recognizer & \textbf{0.7975} & \textbf{85.57\%} & \textbf{0.9847} \\
\bottomrule
\end{tabular}
}
\caption{
Held-out evaluation of the distilled critical-step recognizer on correct
GLM-4.7 trajectories from the BCP held-out split. The original full
trajectories have an origin recall of 0.8229. The student recognizer achieves
teacher-level performance with a single full-trajectory inference pass.
}
\label{tab:heldout_recognizer_eval}
\end{table*}

As shown in Table~\ref{tab:heldout_recognizer_eval}, the distilled recognizer
matches the backward teacher on held-out trajectories, slightly surpassing
the teacher reference on extract recall, evidence coverage accuracy, and step
hit. This indicates that the student can reliably perceive evidence-critical
steps from unseen BCP trajectories with a single inference pass. Together with
the annotation strategy analysis in Table~\ref{tab:annotation_strategy_analysis}, these results support our design of using backward teacher annotation for
offline supervision construction and distilling it into an efficient
single-pass critical-step recognizer.

\subsection{A.3 Sensitivity Analysis of $\alpha$}
\label{app:alpha}

The coefficient $\alpha$ in the efficiency-aware reward
$R^{\mathrm{crit}}_i = K_i/(K_i+\alpha T_i)$ controls how strongly
redundant interaction steps are penalized relative to evidence-critical
ones. A larger $\alpha$ shrinks the reward whenever redundant steps
$T_i$ dominate, pushing the policy toward trajectories with a higher
critical-step ratio, whereas a small $\alpha$ leaves redundant
exploration almost unpenalized. Since this single coefficient governs
the accuracy--efficiency balance that is central to CRISP, we study how
it shapes the learned search behavior. We vary
$\alpha\in\{0.5, 0.7, 1.0\}$ and evaluate on HLE-Verified, comparing all
variants at the same number of optimization steps for a fair comparison.
Table~\ref{tab:alpha} reports answer accuracy, failure rate, and average
interaction turns; the $\alpha=0.7$ setting corresponds to the CRISP
configuration used throughout the main experiments.

The results reveal a non-monotonic trade-off rather than a
``smaller-is-always-shorter'' trend. With a weak penalty ($\alpha=0.5$),
redundant exploration is largely left unpruned: the policy still takes
many interaction turns on average, yet attains the lowest accuracy and
the highest failure rate. Because redundant steps are barely discouraged,
the reward provides little pressure to reallocate interactions toward
evidence-critical ones, so the policy retains noisy trajectories that
inflate both cost and failures. With a strong penalty ($\alpha=1.0$),
interaction cost is minimized and the failure rate is lowest, but accuracy
drops: penalizing redundancy too aggressively begins to suppress steps
that gather necessary evidence, trading answer quality for brevity.
Setting $\alpha=0.7$ attains the best accuracy while already reducing
mean turns to nearly the level of the $\alpha=1.0$ setting. In other
words, most of the efficiency gain is realized once redundant steps are
moderately penalized, and pushing $\alpha$ higher buys only a marginal
turn reduction at a clear accuracy cost.

This pattern is consistent with the central claim of the main paper:
the goal is not to act less, but to preserve evidence-critical
interactions while pruning redundant ones. A moderate $\alpha$ removes
redundancy without eroding the evidence chain, which is exactly why the
$\alpha=0.7$ CRISP policy in our main experiments reduces average turns without
sacrificing final-answer accuracy. We therefore adopt $\alpha=0.7$ in
all main experiments, and note that CRISP is robust to this choice: all
three settings improve efficiency over Vanilla RL, and the accuracy
varies within a narrow range across the studied settings.

\begin{table}[t]
\centering
\small
\setlength{\tabcolsep}{14pt}
\renewcommand{\arraystretch}{1.15}
\begin{tabular}{cccc}
\toprule
$\alpha$ & Acc$\uparrow$ & Fail Rate$\downarrow$ & Avg. Turns$\downarrow$ \\
\midrule
$0.5$          & $31.49$          & $14.25$          & $9.18$ \\
$\mathbf{0.7}$ & $\mathbf{33.41}$ & $12.10$          & $6.58$ \\
$1.0$          & $32.18$          & $\mathbf{11.49}$ & $\mathbf{6.21}$ \\
\bottomrule
\end{tabular}
\caption{Sensitivity of the redundancy penalty $\alpha$ on
HLE-Verified. All variants are trained from the same backbone and
compared at the same number of optimization steps. Acc is final-answer
accuracy, Fail Rate is the percentage of queries with no valid final
response, and Avg.\ Turns is the mean number of tool-interaction turns
per query. $\alpha=0.7$ (used in main experiments) achieves the best
accuracy while retaining most of the efficiency gain.}
\label{tab:alpha}
\end{table}

\subsection{A.4 Training Details}
\label{app:training_details}

We provide the training hyperparameters used for the critical-step
recognizer and policy optimization. The critical-step recognizer is
trained for $3$ epochs with a peak learning rate of $2\times10^{-5}$,
$5\%$ warmup, weight decay $0.1$, global batch size $64$, and maximum
sequence length $131{,}072$. For policy optimization, we use GRPO with
rollout group size $G=8$, global batch size $128$, and a constant
learning rate of $1\times10^{-6}$. The answer reward is binary,
$R_i^{\mathrm{ans}}\in\{0,1\}$. For reproducibility, we fix the random seed to 42 in all experiments
involving randomness. All experiments are run on 8 nodes with a total of 64 NVIDIA H800 80GB
GPUs (8 per node).

\paragraph{OTC-GRPO baseline} The OTC-GRPO baseline follows the original
OTC setting. It is trained from the same GLM-4.5-Air-Midtrain backbone
and uses the same GRPO configuration as Vanilla RL and CRISP, differing
only in the reward: it multiplies the answer-correctness reward by a
tool-efficiency term rather than adding an efficiency-aware signal.
Following the original OTC setting, we set the tool-reward coefficient
$\alpha=1$ and the smoothing constant $c$ to the corresponding maximum
number of turns (i.e., $60$ for BrowseComp and $100$ for HLE-Verified,
matching our interaction budgets).

\end{document}